\documentclass[runningheads]{llncs}

\usepackage{multirow}
\usepackage{graphicx}
\usepackage{tabularx}
\usepackage{amsmath}
\usepackage{amsfonts}
\usepackage{algorithm} 
\usepackage{algpseudocode} 
\usepackage{bm}

\usepackage{color,xcolor}
\usepackage{hyperref}
\hypersetup{colorlinks=true,
linkcolor=blue,
citecolor = blue,
urlcolor=black}

\newcounter{ToDo}
\newcounter{gaocomm}
\newcounter{wangcomm}
\newcounter{Note1}
\definecolor{blue-violet}{rgb}{0.54, 0.17, 0.89}
\definecolor{mygreen}{rgb}{0.0, 0.5, 0.0}
\definecolor{awesome}{rgb}{1.0, 0.13, 0.32}
\definecolor{bostonuniversityred}{rgb}{1.0, 0.0, 0.0}

\begin{document}
%
\title{How Neural Processes\\ Improve Graph Link Prediction}
%
%
\author{Huidong Liang \and Junbin Gao}
\authorrunning{H. Liang and J. Gao}
%
\institute{Discipline of Business Analytics, The University of Sydney Business School\\
The University of Sydney, Sydney, NSW 2006, Australia\\
\email{\href{mailto:hlia0714@uni.sydney.edu.au}{hlia0714@uni.sydney.edu.au}, \href{mailto:junbin.gao@sydney.edu.au}{junbin.gao@sydney.edu.au}}}
\maketitle              
\begin{abstract}
Link prediction is a fundamental problem in graph data analysis. While the majority of the literature focuses on transductive link prediction that requires all the graph nodes and majority of links in training, inductive link prediction, which only uses a proportion of the nodes and their links in training, is a more challenging problem in various real-world applications. In this paper, we propose a meta-learning approach with graph neural networks for link prediction: Neural Processes for Graph Neural Networks (NPGNN), which can perform both transductive and inductive learning tasks, and adapt to patterns in a large new graph after training with a small subgraph. Experiments on real-world graphs are conducted to validate our model, where the results suggest that our model achieves stronger performance compared to other state-of-art models, and meanwhile generalizes well when training on a small subgraph.
\end{abstract}
\section{Introduction}
Graph, consisting of a set of nodes and links, is a common but special data structure in our daily life. With the advancement in machine learning, designing algorithms for problems with graph-type dataset has been successful in many real-world applications~\cite{wu2020comprehensive}. For example, link prediction is one of the important tasks in graph machine learning, in which the goal is to predict some unknown links in a graph given other links and nodes~\cite{liben2007link}. Relevant applications of link prediction involve many areas: in recommendation systems such as friend recommendation~\cite{Adamic2003}, movie recommendation~\cite{koren2009matrix}, and citation recommendation for academic papers~\cite{bhagavatula2018content}; in knowledge discovery in databases (KDD) such as knowledge graph completion~\cite{nickel2015review} and social network analysis~\cite{xu2019link}; and in health science research such as drug-target interaction~\cite{Lu2017} and metabolic network reconstruction~\cite{oyetunde2017boostgapfill}. 

\begin{figure}[t]
\includegraphics[width=\textwidth]{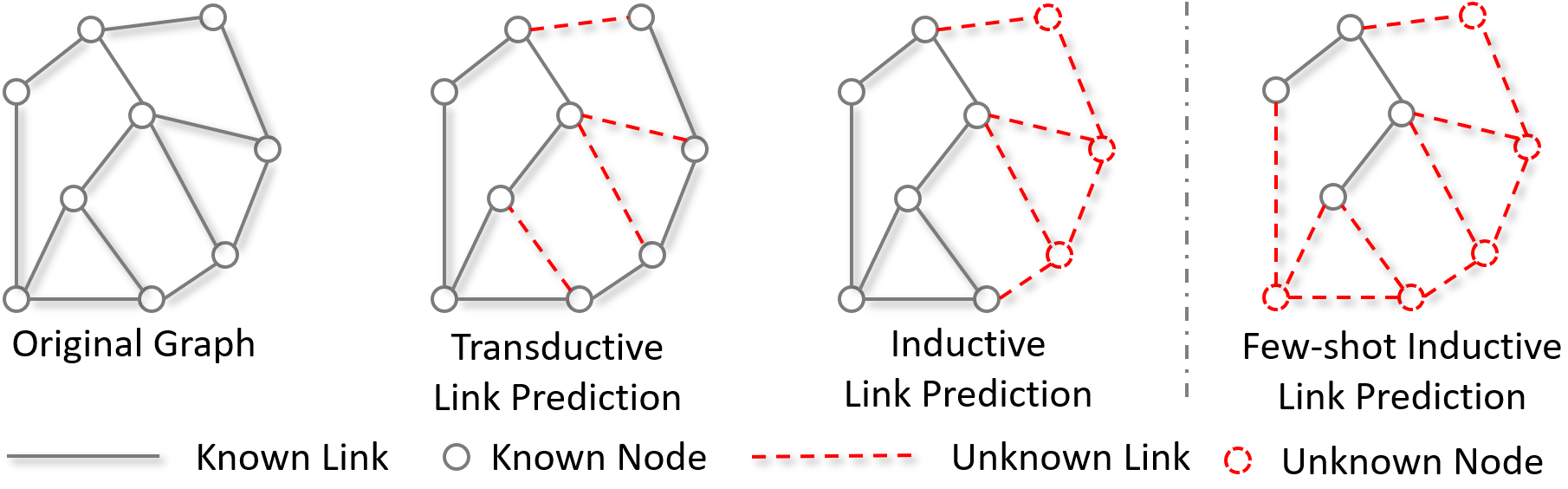}
\caption{Transductive, Inductive, and Few-shot Inductive Link Predictions, with a common goal of predicting unknown links.} \label{fig1}
\end{figure}

Generally, there are two kinds of link prediction tasks: transductive link prediction and inductive link prediction, as illustrated in Fig.~\ref{fig1}. For transductive link prediction, all the nodes information with the majority of links are known in training, and the goal is to predict the unknown links in the entire graph. Whereas for inductive link prediction, a small proportion of nodes are not seen when building up the model, and the remaining nodes with their corresponding links information are used for training. At prediction time, this small proportion of nodes will join the graph, and the goal is to infer the unknown links in the entire new graph. 

Currently, the majority of the literature focuses on transductive link prediction, represented by embedding-based approaches such as DeepWalk (DW) \cite{perozzi2014deepwalk} and Spectral Clustering (SC) \cite{tang2011leveraging}, in which dimensionality reduction techniques are utilized to generate a low-dimensional vector representation for each node's high-dimensional feature in the graph, such that useful information can be exploited efficiently. Another approach with graph embedding, Variational Graph Autoencoders (VGAE) \cite{kipf2016variational}, also considers neighbours' information when generating latent representations for nodes via Graph Convolutional Networks (GCN) \cite{kipf2016semi}, and has shown strong performance on various datasets for link prediction. Nevertheless, in many real-world problems, the size of the graph is growing over time, which requires generating embeddings for new nodes (such as new customers and products in a recommendation system) based on the existing graph and hence performing inductive link prediction. GraphSAGE \cite{hamilton2017inductive}, which generates embeddings via sampling and aggregating features in a node’s local neighbours, provides a way for such tasks, but it assumes the knowledge of some connection between new nodes and existing nodes.

One of the limitations of the above approaches is that they generally assume a relatively large proportion of information to be given during training, and the goal is to predict a small proportion of unknown links~\cite{bose2019meta}. However, what if we only have access to limited information about the graph in training time, and our goal is to predict a large proportion of the links? For example, in recommendation system, if the company is at a start-up stage, in which we only have the information about a small group of customers and products, such as the features of customers and products, and customers' preference towards products (links). As the company is growing fast, a large group of customers and products join the network, but the preference for the new customers towards the products is unknown. In this case, how can we infer the links in the bigger network based on the patterns learned from a smaller graph? We define this type of task as few-shot inductive link prediction, as shown in Fig.~\ref{fig1}.

In this paper, we propose Neural Processes for Graph Neural Networks (NPGNN), in which we implement Neural Processes~\cite{garnelo2018neural} that carries a meta-learning characteristic under graph settings, to perform transductive, inductive and few-shot inductive link predictions. We will first start with introducing Neural Processes and Variational Graph Autoencoders in the Related Work section, and then discuss how we incorporate NP with graph neural networks in our proposed method. Lastly, in the experiment section, we test our model for both transductive and inductive link prediction tasks on three popular citation networks and compare the performance with some other state-of-art models. In addition, we also discuss a few-shot inductive link prediction scenario when only using a small proportion of the graph in training, to predict a large number of unknown links after observing a new set of nodes.

\section{Related work}
\subsection{Variational Autoencoders}
\subsubsection{Model.} Variational Autoencoders (VAE)~\cite{kingma2013auto} is an unsupervised approach for dimensionality reduction in Euclidean space. It aims to generate low-dimensional latent embeddings $\bf z$ with useful information from the original high-dimensional features $\bf x$ via a neural network encoder, such that only a little information is lost when reconstructing this latent representation $\bf z$ back to the high dimensional features $\bf x$ through another neural network decoder. Instead of directly encoding the features into a low-dimensional representation, VAE assumes a latent distribution for $\bf z \sim \mathcal{N}({\bm \mu, \bm \sigma^2})$, and use the encoder to parameterise $\bm \mu$ and $\bm \sigma$. To recreate the original features $\bf x$, the model samples a $\bf z$ from the latent distribution, and send it to the decoder network to reconstruct $\bf x$.

\subsubsection{Inference.} The learning process for VAE is achieved by \textit{Variational Inference}, in which the goal is to minimize the Kullback-Leibler (\textit{KL}) Divergence between the variational distribution $q({\bf z|x})$ and the true posterior distribution $p({\bf z|x})$: 
\[ {\bf KL}\big( q({\bf z|x}) || p({\bf z|x})  \big ) = \mathbb{E}_{q({\bf z|x})}\Bigg [ \log \frac{q({\bf z|x})}{p({\bf z|x})} \Bigg ]. \]

As the above expression also contains the intractable posterior $p({\bf z|x})$ and is problematic, we can rewrite the expression for the \textit{KL} Divergence term as: 
\[{\bf KL}\big( q({\bf z|x}) || p({\bf z|x})  \big ) = \log p({\bf x}) - \mathcal{L},\]
where $\mathcal{L} = \mathbb{E}_{q({\bf z|x})}\Big[ \log  \frac{p({\bf x,z})}{q({\bf z|x})}\Big]$ is the variational lower bound. By Jensen's Inequality ${\bf KL}\big( q({\bf z|x}) || p({\bf z|x})  \big )$ is non-negative, and it is easy to show
$\log p({\bf x}) \geq \mathcal{L}$.

Then, minimizing ${\bf KL}\big( q({\bf z|x}) || p({\bf z|x})  \big )$ is effectively maximizing the variational lower bound $\mathcal{L}$, which can be further re-expressed as: 
\begin{align}
    \mathcal{L} = \mathbb{E}_{q({\bf z|x})}\Big[ \log p({\bf x|z})\Big] - \text{\bfseries KL} \Big( q({\bf z|x})|| p({\bf z}) \Big),\label{vae L}
\end{align} 
where $p({\bf x|z})$ is parameterised by the decoder, $q({\bf z|x})$ is the variational distribution that parameterised by the encoder, and $p({\bf z})$ is the prior distribution for $\bf z$, which is manually selected such as a standard Gaussian distribution. 

To optimize the model, gradients for the parameters need to be computed in backpropagation. For the second \textit{KL} divergence term on RHS, we can compute the gradients analytically. However, for the first expectation term, since $\bf z$ is sampled from a distribution, we need to use the \textit{reparameterization trick} to replace the sampling procedure by a function that contains the parameters in the model, and then estimate the expectation via Monte Carlo methods:
\begin{align}
    \mathbb{E}_{q({\bf z|x})}[\log p({\bf x|z})]  \simeq & \frac{1}{L}  \sum_{l = 1}^L \log p({\bf x|}{\bf z}^{(\ell)}), \label{vae mc}\\
    {\bf z}^{(\ell)} = {\bm \mu} + {\bm \sigma} \epsilon^{(\ell)}, \hspace{0.25 cm} &\text{with} \hspace{0.25 cm} \epsilon^{(\ell)} \sim \mathcal{N}(0,1) \label{vae rp}.
\end{align}

\subsection{Neural Processes}
 \subsubsection{Model.} Neural Processes (NP)~\cite{garnelo2018neural} is a meta-learning approach that aims to learn a way of how to learn new patterns, that is, after training with several tasks (e.g. predicting cats and birds pictures), at prediction time, the model will be asked to predict some new tasks (e.g. predict dogs), given a small sample from the new task.

In training, NP first learns a low-dimensional representation $\bf r_c$ for each data pair $\{ {\bf x}_{c_i}, {\bf y}_{c_i} \}$ in a random Context dataset $C$ (e.g. contains some cats and birds pictures) by neural networks, then aggregate (for example, average) these representations to form a global representation $\bf r$. Similar to VAE, NP also introduces a latent distribution $\bf z \sim \mathcal{N}\big(\bm \mu({\bf r}),\bm \sigma({\bf r})\big)$, in which $\bm \mu({\bf r})$ and $\sigma^2({\bf r})$ are from aggregation and parameterised by neural networks. We can also regard $\bf z$ to be a distribution over functions that follows a \textit{Gaussian Process}. Finally, we use a sampled $\bf z$, together with features ${\bf x}_T$ in a new Target dataset $T$ (e.g. contains some other cats and birds pictures), to predict the respond variable $y_T$ in $T$. 

At prediction time, the model will be given a new Context set from a different task (e.g. some pictures of dogs), and it will generate a latent representation $\bf z$ that contains the information of the ``patterns'' in the new task, and together with the features from a new Target set, to predict the response variable in the new Target set (e.g. some other dog pictures).

\subsubsection{Inference.} NP uses \textit{Variational Inference} with similar settings in VAE \eqref{vae L}, in which the goal is to maximize the variational lower bound $\mathcal{L}$:
\begin{align}
    \mathcal{L} = \mathbb{E}_{q({\bf z|D})}\Big[ \log p({\bf y}_T |{\bf x}_T,{\bf z})\Big] - \text{\bfseries KL} \Big( q({\bf z}|D)|| q({\bf z}|C) \Big), \label{NP L}
\end{align} 
where $D = C \cup T$ is used for generating more informative $\bf z$ during inference, which is parameterised by the encoder. And instead of adopting a standard prior, NP chooses $q({\bf z}|C)$ that is encoded by the random Context C as the prior $p({\bf z})$. As such, the inference forces the information inferred from a random context set to be close to the information inferred from the overall dataset.

To optimize $\mathcal{L}$, NP also implements Monte Carlo methods with \textit{reparameterization trick} to estimate the gradient in the first RHS expectation term in equation \eqref{NP L}, with similar approach as equations \eqref{vae mc} and \eqref{vae rp} in VAE.

Although one previous work \cite{carr2019graph} discussed Conditional Neural Processes \cite{garnelo2018conditional} (which does not involve the latent variable $\bf z$) on graph edge imputation, the implementation of Neural Process to generate graph embedding for link prediction is still unexplored, and will be discussed in our work.

\subsection{Variational Graph Autoencoders}
Variational Graph Autoencoders (VGAE)~\cite{kipf2016variational} is an implementation of Variational Autoencoders (VAE)~\cite{kingma2013auto} on graph-type dataset with Graph Convolutional Networks (GCN)~\cite{kipf2016semi}, which can generate graph embeddings for every node by considering the node's neighbours information (features). Similar to VAE, it chooses two GCNs as the encoder to generate latent graph embeddings from latent Gaussian distributions, which are then sent to an inner-product decoder to predict links in the graph. The learning for VGAE is also carried out by \textit{Variational Inference} and is consistent with VAE.

\section{Proposed Model: NPGNN}
\subsection{Setup and Framework}
First we define an undirected graph ${\bf G} =({\bf V, A})$, where $\bf V$ is the set of nodes with features ${\bf x}_i \in \bf X$ corresponding to each node ${\bf v}_i \in \bf V$, and the adjacency matrix $\bf A$ with ${\bf A}_{ij} = 1$ if there is a link between ${\bf v}_i$ and ${\bf v}_j$, and ${\bf A}_{ij} = 0$ otherwise. Then we randomly select a subset of nodes $\bf V_C \in V$ with its related features $\bf X_C \in X$ and adjacency matrix $\bf A_C$ to construct a context subgraph $\bf G_C$. We assume there are $n$ nodes in the complete graph $\bf G$, and the first $m$ nodes are the context nodes $\bf V_C$. Our goal is to model the adjacency matrix $\bf A$ for the complete graph $\bf G$ conditional on the context subgraph $\bf G_C$.

\begin{figure}[ht]
\begin{center}\includegraphics[width=0.95\textwidth]{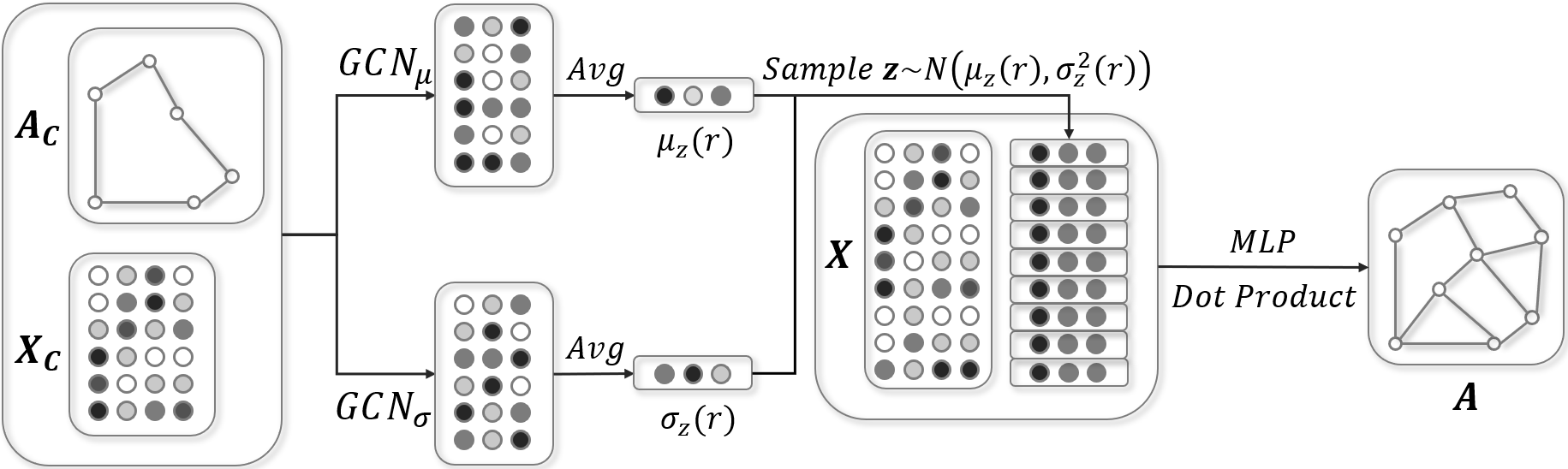}\end{center}
\caption{Framework of Neural Processes for Graph Neural Networks} \label{fig2}
\end{figure}

Fig.~\ref{fig2} illustrates the framework in our model, which starts with two {\bfseries Graph Convolutional Encoders} that encode each node ${\bf v}_i \in \bf V_C$ with feature ${\bf x}_i \in \bf X_C$ and the adjacency matrix $\bf A_C$ to a latent representation ${\bf r}_{{\bf c}_i}$ under multi-variate Gaussian distribution with mean ${\bm \mu}_{{\bm c}_i}$ and variance ${\bm \sigma}^2_{{\bm c}_i}$. We then {\bfseries Aggregate} (average) the latent representations to obtain a global representation $\bf r$ that parameterises the latent distribution $q({\bf z}|{\bf A_C, X_C}) = \mathcal{N} ({\bm \mu}_{\bf z}({\bf r}), \text{diag} ({\bm \sigma}^2_{\bf z}( {\bf r} )))$. Finally, a sampled $\bf z$ is concatenated to each feature ${\bf x}_i \in \bf X$, and together they are sent to a {\bfseries Multilayer Perceptron with Inner Product Decoder} to reconstruct the similarity matrix of $\bf A$ for the complete graph $\bf G$.

\subsection{Graph Convolutional Encoder}
We assume a $d$-dimensional latent representation ${\bf r}_{{\bf c}_i} \in \mathbb{R}^d$ for each context node ${\bf v}_i \in \bf V_C$ under Multi-variate Gaussian Distribution $\mathcal{N}({\bm \mu_{{\bf c}_i}}, \, \text{diag} ({\bm \sigma}^2_{{\bf c}_i})\, )$, and our model uses two two-layer GCNs to encode such distribution for each latent representation ${\bf r}_{{\bf c}_i}$:
\begin{align}
    {\bm \mu_{\bf c}} &= \text{ReLU}( \Bar{\bf A}_{\bf C} \, \text{ReLU}(\Bar{\bf A}_{\bf C} {\bf X_C} {\bf W}_1) {\bf W}_{\bm \mu} ) \label{eq 1},\\
    \log {\bm \sigma_{\bf c}} &= \text{ReLU}( \Bar{\bf A}_{\bf C} \, \text{ReLU}(\Bar{\bf A}_{\bf C} {\bf X_C} {\bf W}_1) {\bf W}_{\bm \sigma} ) \label{eq 2},
\end{align}
where $\Bar{\bf A}_{\bf C} = \Tilde{\bf D}_{\bf C}^{-\frac{1}{2}} \Tilde{\bf A}_{\bf C} \Tilde{\bf D}_{\bf C}^{-\frac{1}{2}}$ is the adjacency matrix for subgraph $\bf G_C$ after convolution, $ \Tilde{{\bf D}}_{{\bf C}_{ii}} = \sum_j \Tilde{\bf A}_{{\bf C}_{ij}}$ is the degree matrix of $\Tilde{\bf A}_{\bf C}$, and $\Tilde{\bf A}_{\bf C} = \bf A_C + I$. Both function \eqref{eq 1} and function \eqref{eq 2} share the same parameters ${\bf W}_1$ in their first layer, and use $\text{ReLU}(t) = \max(0,\,t)$ as the activation function for both layers. The output $\bm \mu_{\bf c}$ in \eqref{eq 1} is a $m \times d$ matrix of mean vectors ${\bm \mu}_{{\bf c}_i}$, and similarly log $\bm \sigma_{\bf c}$ in \eqref{eq 2} is the matrix of standard deviation vectors log ${\bm \sigma}_{{\bf c}_i}$.

\subsection{Aggregation and Latent Embedding}

We then aggregate the latent representation ${{\bf r}_{{\bf c}_i}}$ by averaging to obtain a global representation $\bf r$, which can be used to parameterise the latent probability distribution $q({\bf z}|{\bf A_C, X_C}) = \mathcal{N} ({\bm \mu}_{\bf z}({\bf r}), {\bm \sigma}_{\bf z}^2({\bf r}))$:
\begin{align}
    {\bm \mu}_{\bf z} ({\bf r}) = \frac{1}{m} \sum_{i=1}^{m}{\bm \mu}_{{\bf c}_i}, \hspace{1cm} \log {\bm \sigma}_{\bf z} ({\bf r}) = \frac{1}{m} \sum_{i=1}^{m} \log {\bm \sigma}_{{\bf c}_i}. \label{eq 3}
\end{align}
Similar to Neural Processes, we can view this latent distribution $q$ as a \textit{Gaussian Process} that defines a distribution over many functions, where each of them defines a mapping from the node features $\bf X_C$ in a particular random context subgraph $\bf G_C$ to the corresponding latent representation $\bf r_c$. And then after aggregation over $\bf r_c$, we can define a latent global representation $\bf z$ as a \textit{Gaussian Process}: ${\bf z} \sim \mathcal{G}\mathcal{P} \big( {\bm \mu}_{\bf z}({\bf r}), {\bm \sigma}_{\bf z}^2({\bf r}) \big)$, in which the mean function and kernel function are parameterised by the encoder.

\subsection{Multilayer Perceptron with Inner Product Decoder}
After aggregation, we combine a sampled $\bf z$ with every feature ${\bf x}_i \in \bf X$ as $\Tilde{\bf x}_i \in \Tilde{\bf X}$, where $\Tilde{\bf x}_i = [{\bf x}_i^\top \; {\bf z}^\top ]^\top$. As such, information from context subgraph flows to the complete graph via this latent space $\bf z$, and then we send them to a 2-layer MLP decoder to produce the latent embedding $\bf U$:
\begin{align*}
    {\bf U} = \sigma({\bf W}_3\sigma({\bf W}_2\Tilde{\bf X} + {\bf b}_1) + {\bf b}_2),
\end{align*}
where $\bf U$ is the matrix of latent embedding vectors ${\bf u}_i$ for each node ${\bf v}_i \in \bf V$ in the complete graph $\bf G$, and $\sigma (t) = 1/(1 + \exp(- t))$ is the logistic sigmoid activation function.

Finally, we take inner product for each ${\bf u}_i$, and use sigmoid function to calculate the probability of edge  existence between two nodes ${\bf v}_i, {\bf v}_j \in \bf V$, and the likelihood $p(\bf A|X,z)$:
\begin{align}
    p({\bf A|X,z}) = \prod_{i=1}^n \prod_{j=1}^n p({\bf A}_{ij}|{\bf u}_i,{\bf u}_j), \hspace{0.25 cm} \text{with} \hspace{0.25 cm} p({\bf A}_{ij} = 1|{\bf u}_i, {\bf u}_j) = \sigma({\bf u}_i^\top {\bf u}_j). \label{eq 5}
\end{align}
\subsection{Inference and Learning} 
\begin{figure}[t]
\begin{center}
\includegraphics[width=0.95\textwidth]{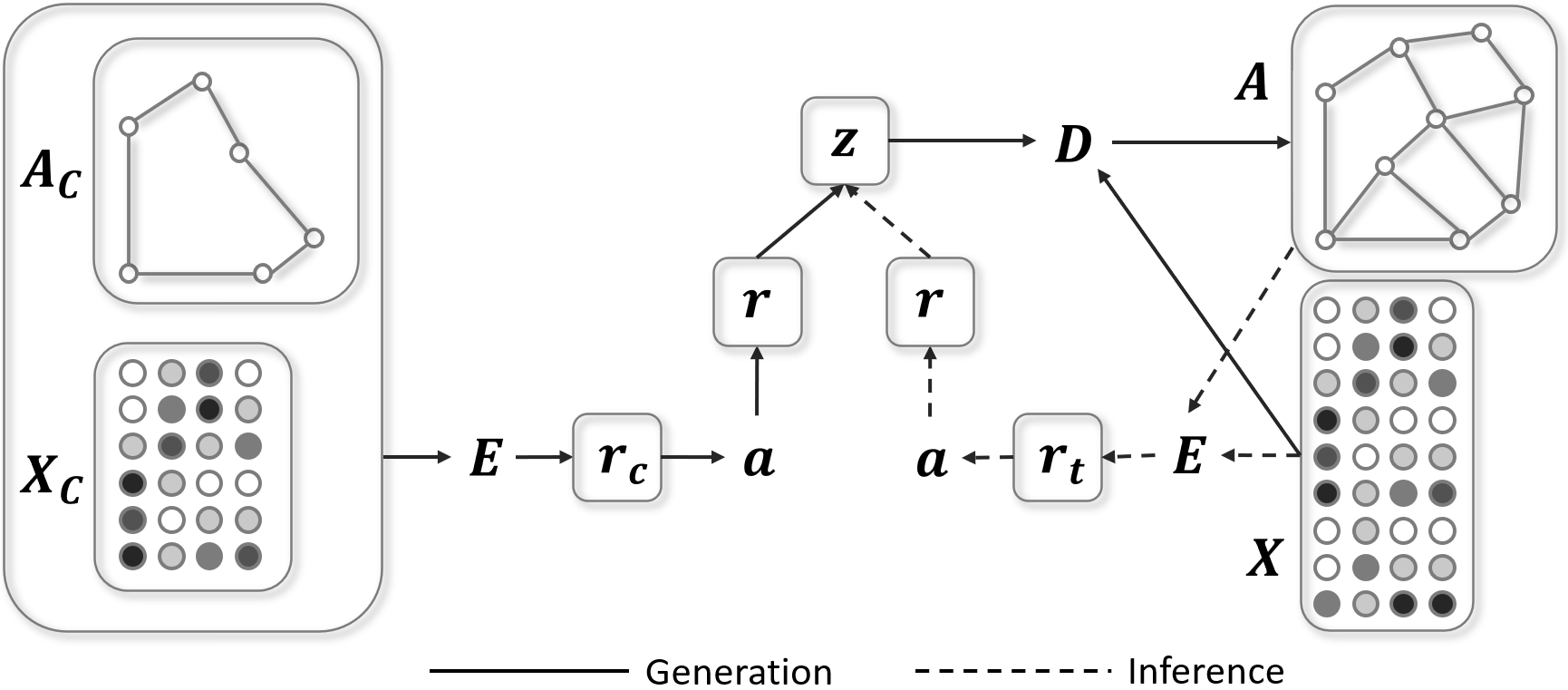}
\end{center}
\caption{Inference of Neural Processes for Graph Neural Networks. Explanation for notations in the figure: E - encoder, D - decoder and a - aggregator.} \label{fig3}
\end{figure}

Inference for our model is carried out by \textit{Variational Inference}, and is demonstrated in Fig.~\ref{fig3}.

Similar to equation \eqref{NP L} in NP, we can express the variational lower bound $\mathcal{L}$ for NPGNN as:
\begin{align}
    \mathcal{L} = \mathbb{E}_{q({\bf z|A,X})}[\log p({\bf A|X,z})] - \text{\bfseries KL}[ q({\bf z|A,X}) || q({\bf z|A_C,X_C}) ], \label{eq 9}
\end{align}
where $p({\bf A|X,z})$ is parameterised by our decoder model in equation \eqref{eq 5}, and $q({\bf z|A,X})$ is parameterised by the encoder based on the complete graph $\bf G$ as:
\begin{align*}
    q({\bf z|A, X}) = \mathcal{N}\big({\bm \mu}_{\bf z}({\bf r}), \text{diag} ({\bm \sigma}^2_{\bf z}({\bf r})) \big),
\end{align*}
with $\bf r$ aggregated from $\bf r_t$ encoded by $\bf A$ and $\bf X$ in the complete graph $\bf G$ (similar to equation \eqref{eq 3}, where we use $\bf A_C$ and $\bf X_C$ from the context subgraph $\bf G_C$ to encode $\bf z$ in stead). Here the reason that we choose $\bf G$ in forward pass is because at prediction time, we will be predicting the links in a larger graph conditional on the information of the ``complete" graph $\bf G$ we have in training, thus $q({\bf z|A, X})$ will lead to a more informative $\bf z$. For the prior, we choose $q({\bf z|A_C,X_C})$ encoded from the random context subgraph $\bf G_C$. 

Therefore, maximizing the variational lower bound $\mathcal{L}$ is now effectively maximizing the expectation of the conditional log likelihood, and meanwhile forcing the latent distribution inferred from different random subgraphs, to be close to the latent distribution inferred from the complete graph. 

To optimize $\mathcal{L}$ in equation \eqref{eq 9}, similar to NP and VAE, we use Mote Carlo methods (equation \eqref{eq 10}) to estimate the expectation in the first RHS term by sampling $\{{{\bf z}^{(1)},...,{\bf z}^{(L)}}\}$ from $\mathcal{N} ({\bm \mu}_{\bf z}({\bf r}), {\bm \sigma}_{\bf z}^2({\bf r}))$ with \textit{reparameterization trick} (equation \eqref{eq 11}), which will lead to a closed-form gradient in backpropagation:
\begin{align}
    \mathbb{E}_{q({\bf z|A,X})}[\log p({\bf A|X,z})]  \simeq & \frac{1}{L}  \sum_{l = 1}^L \log p({\bf A|X},{\bf z}^{(l)}), \label{eq 10}\\
    {\bf z}^{(l)} = {\bm \mu}_{\bf z}({\bf r}) + {\bm \sigma}_{\bf z}({\bf r}) \epsilon^{(l)}, \hspace{0.25 cm} &\text{with} \hspace{0.25 cm} \epsilon^{(l)} \sim \mathcal{N}(0,1). \label{eq 11}
\end{align}

\begin{algorithm}[t]
	\caption{Neural Processes for Graph Neural Networks} \label{algo1}
	\textbf{Input}: Complete Graph ${\bf G} = ({\bf V,A})$ with features $\bf X$
	
	\hspace{1cm} Encoder function $Encode(\cdot,\cdot)$; Decoder function $Decode(\cdot,\cdot)$
	
	\hspace{1cm} Aggregation function $Aggregate(\cdot)$ 
	
	\hspace{1cm} Total iteration $T$; Number of Monte Carlo estimates $L$
	
    \textbf{Output}: Optimized $Encode(\cdot,\cdot)$ and $Decode(\cdot,\cdot)$.
	\begin{algorithmic}[1]
    \For {iteration $= 0, 1, 2, ..., T$}
        \State Generate a random context subgraph ${\bf G_C} = ({\bf V_C, A_C})$ with features $\bf X_C$
		\State $\bf r_c$ $\leftarrow$  $Encode({\bf A_C, X_C})$ \hspace{1cm} $\bf r_t$ $\leftarrow$  $Encode({\bf A, X})$
		\State $\bf r'$ $\leftarrow$ $Aggregate$($\bf r_c$) \hspace{1.55cm} $\bf r$ $\leftarrow$ $Aggregate$($\bf r_t$)
		\State Compute $q({\bf z|A_C,X_C}) = \mathcal{N}\big({\bm \mu}_{\bf z}({\bf r'}), \text{diag} ({\bm \sigma}^2_{\bf z}({\bf r'})) \big) $
		\State Compute $q({\bf z|A,X}) = \mathcal{N}\big({\bm \mu}_{\bf z}({\bf r}), \text{diag} ({\bm \sigma}^2_{\bf z}({\bf r})) \big) $	
		\State Sample $L$ $\bf z$ $\sim q({\bf z|A,X}) $
		\State $p({\bf A|X,z}) $ $\leftarrow$ $Decode({\bf X,z})$
		\State Compute $\nabla \mathcal{L}$ for $\mathcal{L} = \frac{1}{L}  \sum_{l = 1}^L \log p({\bf A|X},{\bf z}^{(l)}) - \text{\bfseries KL}[ q({\bf z|A,X}) || q({\bf z|A_C,X_C}) ]$
		\State Update $Encode(\cdot,\cdot)$ and $Decode(\cdot,\cdot)$ when optimizing $\mathcal{L}$ with $\nabla \mathcal{L}$
	\EndFor\\
	\Return $Encode(\cdot,\cdot)$ and $Decode(\cdot,\cdot)$
	\end{algorithmic} 
\end{algorithm}

As such, we can optimize our model by standard optimization tools, with an algorithm for learning summarised in Algorithm~\ref{algo1}.

\section{Experiments}
\subsection{Experimental Set-up}
\subsubsection{Dataset, Metrics and Code:} To validate our proposed model, we will be conducting two experiments for transductive link prediction and inductive link prediction. We also consider a situation where only a small proportion of graph is known for training after inductive experiment, and proceed to a few-shot inductive link prediction scenario. For each experiment, we will test our model on two regular-size citation networks: Cora (2,708 nodes with 1,433-dimensional features, and 5,429 links) and Citeseer (3,327 nodes with 3,703-dimensional features, and 4,732 links), and one large citation network PubMed (19,717 nodes with 500-dimensional features, and 44,338 links). 


We measure our model's performance by AUC score (the
Area Under a receiver operating characteristic Curve) and AP score (Average Precision) after 10 runs with different random seeds, and report the mean scores and their standard errors. All the code for replicating the following results can be found at \url{https://github.com/LeonResearch/NPGNN}.
\subsubsection{Baseline Models:}
We compare our model's performance against other embedding based state-of-art methods. For transductive link prediction, we compare our model with Spectral Clustering (SC)~\cite{tang2011leveraging}, DeepWalk (DW)~\cite{perozzi2014deepwalk}, and Variational Graph Autoencoders (VGAE)~\cite{kipf2016variational}. For inductive link prediction, we compare our model with VGAE, after which we analyze the results for our proposed method when training on three different proportions of the complete graph.

\subsubsection{Hyper-parameters Settings:} For both experiments, we use two 32-neuron hidden layers in the encoder, and a 2-hidden-layer MLP with 64 neurons and 32 neurons respectively. We train our model 500 iterations on Cora and Citeseer by Adam algorithm~\cite{kingma2014adam} with a learning rate of 0.01 and $\beta = [0.9, \, 0.009]$, and we initialize weights as described in \cite{glorot2010understanding}. Since the PubMed dataset is relatively large, we train our model for 4,000 iterations by Adam under the same settings. For other baseline models, we maintain the settings in the corresponding papers.

\subsection{Transductive Experiment} 
We maintain the setting used in VGAE~\cite{kipf2016variational} that randomly masks 10\% edges for testing, 5\% edges for validation, and uses the rest edges to construct training adjacency matrix $\bf A_T$ with features $\bf X$ for all nodes during training. When building the context subgraph in our model, we randomly select 10\% training edges to construct the context adjacency matrix $\bf A_C$, and regard features for all nodes $\bf X$ as the context features $\bf X_C$. At prediction, we use training adjacency matrix $\bf A_T$ and features $\bf X$ for all nodes as the context subgraph to predict the complete adjacency matrix $\bf A$.

\begin{table}\centering
\caption{Transductive Link Prediction Results}\label{tab2}
\setlength{\tabcolsep}{1.5 pt}
\begin{tabular}{l |c c c c c c}
\hline
\multirow{2}{*}{\bfseries Method} & \multicolumn{2}{c}{Cora} & \multicolumn{2}{c}{Citeseer} & \multicolumn{2}{c}{PubMed}\\
& AUC & AP & AUC & AP& AUC & AP\\
\hline 
\multirow{2}{*}{SC} & 84.6& 88.5 & 80.5 & 85.0 & 84.2 & 87.8\\
& (0.01) & (0.00) & (0.01) & (0.01) & (0.02) & (0.01)\\
\hline
\multirow{2}{*}{DW}& 83.1    & 85.0    & 80.5    & 83.6    & 84.4    & 84.1   \\
& (0.01) & (0.00) & (0.02) & (0.01) & (0.00) & (0.00)\\
\hline
\multirow{2}{*}{VGAE}  & 91.4   & 92.6   & 90.8  & 92.0  & 94.4    & 94.7    \\
& (0.01) & (0.01) & (0.02) & (0.02) & (0.02) & (0.02)\\
\hline
\multirow{2}{*}{NPGNN} & {\bfseries 93.1} &	{\bfseries 94.0} &	{\bfseries 94.0} &    {\bfseries 95.1} &	{\bf 95.3}    &	{\bf 95.2}    \\
& ({\bf 0.004}) & ({\bf 0.004}) & ({\bf 0.002}) & ({\bf 0.002}) & ({\bf 0.001}) & ({\bf 0.001})\\
\hline
\end{tabular}
\end{table}

\noindent The results are summarized in Table~\ref{tab2}. It shows that our model outperforms other baseline models on all three datasets, with a relatively large margin on Cora and Citeseer for both $AUC$ and $AP$ scores, and achieves results with a slight improvement on PubMed compared to VGAE.

\subsection{Inductive Experiment} 
The settings for the inductive experiment is different from the transductive experiment. Here, we randomly select 5\% nodes and use the links adjacent to these nodes for testing (roughly 10\% of all links in Cora, 7\% in Citeseer, and 11\% in PubMed), and use the same method to select 2.5\% nodes and their corresponding links for validation. The rest of the nodes with the links among them are used to build the training adjacency matrix $\bf A_T$, and the features $\bf X_T$ for these nodes  are used in training. When building up our model, we randomly use 10\% of training nodes with their connected edges to construct the context adjacency matrix $\bf A_C$, and again only use the features for that 10\% nodes as context features $\bf X_C$. At prediction, we treat the training adjacency matrix $\bf A_T$ and training features $\bf X_T$ as the context subgraph, then use features $\bf X$ for all nodes to predict the adjacency matrix $\bf A$ for the complete graph. 

\begin{table}[h]
\centering
\caption{Inductive Link Prediction Results}\label{tab3}
\setlength{\tabcolsep}{1.5 pt}
\begin{tabular}{l| c c| c c| c c}
\hline
\multirow{2}{*}{\bfseries Method} & \multicolumn{2}{c|}{Cora} & \multicolumn{2}{c|}{Citeseer} & \multicolumn{2}{c}{PubMed}\\
& AUC & AP & AUC & AP& AUC & AP\\
\hline 
\multirow{2}{*}{VGAE} & 77.6 & 73.2 & 82.2 & 79.5 & 84.3 &	80.8 \\
&(0.025)&(0.018)&(0.021)&(0.028)&(0.006)&(0.007)\\
\hline
\multirow{2}{*}{NPGNN} & {\bfseries 85.0} & {\bfseries 85.9} &	{\bfseries 91.0} & {\bfseries 91.8} &  {\bfseries 94.0} & {\bfseries 94.0} \\
&({\bf 0.024})&({\bf 0.024})&({\bf 0.012})&({\bf 0.013})&({\bf 0.001})&({\bf 0.001})\\
\hline
\end{tabular}
\end{table}

\noindent Table~\ref{tab3} summarises the results for inductive link prediction, and it shows our model achieves both higher $AUC$ scores and $AP$ scores on all three datasets by a significant difference from VGAE.

\subsubsection{Few-shot Inductive Experiment:} We also consider a scenario when only a small group of links are seen during training, while at prediction, our goal is to predict the rest of the links. We test our model with three different proportions of graph used in training: (1) 30\% nodes and the links among them (around 10\% total links), (2) 50\% nodes and the links among them (around 25\% total links), and (3) 70\% nodes and the links among them (around 50\% total links). Then, we use the rest (1) 90\% total links, (2) 75\% total links, and (3) 50\% total links for testing. Since PubMed is a large network with around $20k$ nodes and $44k$ links, constructing subgraphs for training is heavily time-consuming on the CPU. As such, we only test our model's few-shot inductive link prediction performance on Cora and Citeseer, with results summarized in Table~\ref{tab4}.

\begin{table}
\centering
\caption{Few-shot Inductive Link Prediction Results}\label{tab4}
\setlength{\tabcolsep}{3 pt}
\begin{tabular}{l| c c| c c}
\hline
\multirow{2}{*}{\bfseries Training Graph Size} & \multicolumn{2}{c|}{Cora} & \multicolumn{2}{c}{Citeseer}\\
& AUC & AP & AUC & AP\\
\hline 
\multirow{2}{*}{30\% nodes (10\% links)} & 74.8  &	76.6 &	  84.0 &  85.6 \\
&(0.005) &(0.007) &(0.006) &(0.006)\\
\hline
\multirow{2}{*}{50\% nodes (25\% links)} & 78.8  &	80.4 &	{ 87.3 } & {  88.7 }   \\
&(0.009) &(0.009) &(0.004) &(0.003)\\
\hline
\multirow{2}{*}{70\% nodes (50\% links)} & 81.9  &	83.3 &	{ 89.4 } & {  90.6 }   \\
&(0.011) &(0.012) & (0.006) &(0.005)\\
\hline
\end{tabular}
\end{table}

\noindent The results suggest that even if NPGNN is trained with 30\% nodes and links among them (around 10\% total links) when predicting the rest 90\% links, our proposed method still achieves a descent performance of 74.8\% $AUC$ and 76.6\% $AP$ on Cora, and
84.0\% $AUC$ score and 85.6\% $AP$ score on Citeseer. This shows NPGNN's generalizability of learning useful embeddings given a small sample size to predict a large proportion of links with unseen nodes.

\section{Conclusion}
In this paper, we introduce a novel approach Neural Processes for Graph Neural Networks that generates a global latent embedding as a distribution over functions on a context subgraph, which can be later used to predict links for both transductive and inductive learning. We also show our proposed model experimentally on different real-world graphs for three types of link prediction, where NPGNN achieves strong performance when comparing with state-of-art models, and also generalizes well on a larger graph when only training on a graph with much smaller size.

%
%
%
\bibliographystyle{splncs04}
\bibliography{ref}
\end{document}